# LiDAR Data Enrichment Using Deep Learning Based on High-Resolution Image: An Approach to Achieve High-Performance LiDAR SLAM Using Low-cost LiDAR

Jiang Yue[1,2], Weisong Wen[1], Jing Han[2], and Li-Ta Hsu[1]
[1]Hong Kong Polytechnic University, Hong Kong
[2]Nanjing University of Science and Technology, China

*Abstract*—LiDAR-based simultaneous localization and mapping (SLAM) algorithms are extensively studied to providing robust and accurate positioning for autonomous driving vehicles (ADV) in the past decades. Satisfactory performance can be obtained using high-grade 3D LiDAR with 64 channels, which can provide dense point clouds. Unfortunately, the high price significantly prevents its extensive commercialization in ADV. The cost-effective 3D LiDAR with 16 channels is a promising replacement. However, only limited and sparse point clouds can be provided by the 16 channels LiDAR, which cannot guarantee sufficient positioning accuracy for ADV in challenging dynamic environments. The high-resolution image from the low-cost camera can provide ample information about the surroundings. However, the explicit depth information is not available from the image. Inspired by the complementariness of 3D LiDAR and camera, this paper proposes to make use of the high-resolution images from a camera to enrich the raw 3D point clouds from the low-cost 16 channels LiDAR based on a state-of-the-art deep learning algorithm. An ERFNet is firstly employed to segment the image with the aid of the raw sparse 3D point clouds. Meanwhile, the sparse convolutional neural network (SCNN) is employed to predict the dense point clouds based on raw sparse 3D point clouds. Then, the predicted dense point clouds are fused with the segmentation outputs from ERFnet using a novel multi-layer convolutional neural network (MCNN) to refine the predicted 3D point clouds. Finally, the enriched point clouds are employed to perform LiDAR SLAM based on the state-of-the-art normal distribution transform (NDT). We tested our approach on the re-edited KITTI datasets: (1) the sparse 3D point clouds are significantly enriched with a mean square error of 1.1m MSE. (2) the map generated from the LiDAR SLAM is denser which includes more details without significant accuracy loss.

## I. Introduction

Accurate positioning is a fundamental part of the realization of safety-critical Level 4 (L4) autonomous driving vehicles (ADV) [1]. The light detection and ranging (LiDAR) is an indispensable sensor for providing positioning at a high frequency by simultaneous localization and mapping (SLAM) [2-4] or map matching [5]. Satisfactory accuracy can be achieved using high-grade 3D LiDAR with 64 channels. Unfortunately, the lasting high price is still the major barrier that prevents its commercialization for ADV. The cost-effective 16 channel LiDAR is a promising replacement of the 64 channel one. However, the positioning accuracy cannot be guaranteed in challenging areas as only limited and sparse point clouds are provided by the low-cost LiDAR. The low-cost camera can provide ample and extra texture information of the surroundings at a high frequency. However, the depth information is not directly available. Therefore, the camera and the 3D LiDAR is, in fact, complementary. Inspired by this, this paper proposes to make use of the high-resolution images from a camera to enrich the raw 3D point clouds from the low-cost 16 channels LiDAR based on a state-of-the-art deep learning algorithm.

## II. Related Work

Scholarly work on point cloud completion is extensive. In this section, however, we focus on reviewing the progress of point cloud enrichment which can further reduce the cost and improve the performance of the LiDAR SLAM algorithm.

To reduce the cost of 3D LiDAR sensors, the straightforward method is to achieve this from the sensor aspect. Numerous efforts are applied to develop low-cost LiDAR such as solid-state LiDAR [6, 7]. Unfortunately, the performance is still not satisfactory and there is still a long way to go.

Alternatively, owning to the dramatic progress of the deep learning technique, numerous work is conducted to enrich the 3D point clouds using deep neural networks (DNN). Typically, the fusion of the spare LiDAR point cloud and the high-resolution image would a good potential solution. For a typical setup [8], the image is more than 10 times denser compared with the LiDAR point cloud, which can be seen in Fig.1. The data enrichment approach is well-known as super-resolution in the field of computer vision [9]. This super-resolution is started with filter research. It draws more attentions when deep learning-based methods are proposed [10, 11]. With the great success of deep learning, it became a common approach to feed the sparse depth maps and high-resolution images to a neural network to train a model. Considering the applications of autonomous driving such as lane-keeping, normally, the lane width would be around 0.1 meters. Since the depth completion is an ill-posed problem [10, 12], the neuronal network can achieve a reasonable result on the training dataset, but the predicted result still far from expectation. To deal with the insufficiency of conventional networks, a sparse convolution network that explicitly considers the location of missing data is proposed to realize super-resolution on sparse depth, achieving a performance of mean absolute error (MAE) about 0.54 meter [13]. The result is the baseline of depth completion on the KITTI [8] depth dataset, and the input is depth only. Compared with traditional algorithms including Markov random field (MRF) and conjugate gradient, the new depth features employed in neuronal networks have significantly reduced the MAE in the depth resolution. Surface normal used as a new depth local representation is proposed to predict the neighborhood pixel depth, it reduced the MAE of result about 0.226 meters [14]. Also, similar to the guided image filter, a pixel is a weighted average of nearby pixels. The weights are inferred from the image by the neuronal network and applied to sparse depth for a high dense depth map [15]. The results show that it reduced the MAE around 0.218 meters, currently, it ranks the first at the KITTI leaderboard (by Dec 2019).

The deep learning has made great progress on the LiDAR enrichment. Unfortunately, the trained models are heavily depending on the dataset. In reality, the captured images are suffered from several unexpected factors, such as heavy traffic, occlusion, high dynamic scene, etc. The most used dataset KITTI is captured in normal scenarios, where the traffic is relatively light comparing to urban areas. The highly urbanized city, such as Hong Kong, the traffic is heavy and containing highly dynamic objects. The depth map enrichment methods have difficulties to deal with the discontinuous of the objects [16]. Actually, the continuous of the depth is common, which is the same as the ordinary image. This is why depth estimation can be naturally formulated as a continuous conditional random field problem [17]. In other words, the depth map is easily reconstructed while the scene is continuous. The problems are normally caused by occlusions and boundaries. Beyond the image, we try to find out a solution to this problem and give a preliminary demonstration.

All in all, the depth completion has drawn significant attention in the deep leaning. Different conventional layers and structures are presented to improve the results. A general confusion is that what kind of information the image and sparse depth could provide. As far as we know, we can hardly find the answer in the present literature. We started with this question, and use the answer to design a new natural network, taking advantage of depth and image, respectively. More clearly, we try to figure out the sparsity of the depth data, due to the compression and enrichment depending on the sparsity. On the other hand, the depth enrichment would be better, if we have a good segmentation result of the image. It is the reason why recent researches focus on the pre-trained model in depth completion [12, 18]. Since the depth is a naturally good source to detect the targets, we analyzed the performance of the target detection with depth input. Based on our analysis results, we designed our neuronal network.

The remainder of this paper is structured as follows. An overview of the proposed method is given in Section III. Section IV presents the proposed methodology before the experimental evaluation is presented in Section V. Finally, the conclusions and future work are drawn in Section VI.

### III. OVERVIEW OF THE PROPOSED METHOD

The overview of the proposed method is shown in Fig. 1. The convolutional network is employed to complete the depth data, simulated 16 channels LiDAR data, which is transformed by the sparse point cloud acquired by 64 channels LiDAR. We model the depth completion as a target segmentation to deal with the discontinuity of the objects. The ERFNet [19] is employed to extract the segmentation result of the image, the input included sparse LiDAR data and image. But the segmentation is heavily dependent on the prior information (pre-trained model) to get a good target segmentation. Since we discovered that depth is better on the detection of the target than the image, we proposed to use the depth and image as target segmentation input. Furthermore, since the LiDAR point cloud is sparse, it is difficult for the conventional kernel to distinguish between observed inputs and those being invalid [13]. Thus, the sparsity invariant CNN (SCNN) is employed to deal with the sparse input, namely LiDAR data. After that, a four convolutional layers branch is proposed to smooth the segmentation and the predicated LiDAR result. Finally, the enriched 3D point clouds are employed to perform the LiDAR NDT based SLAM.

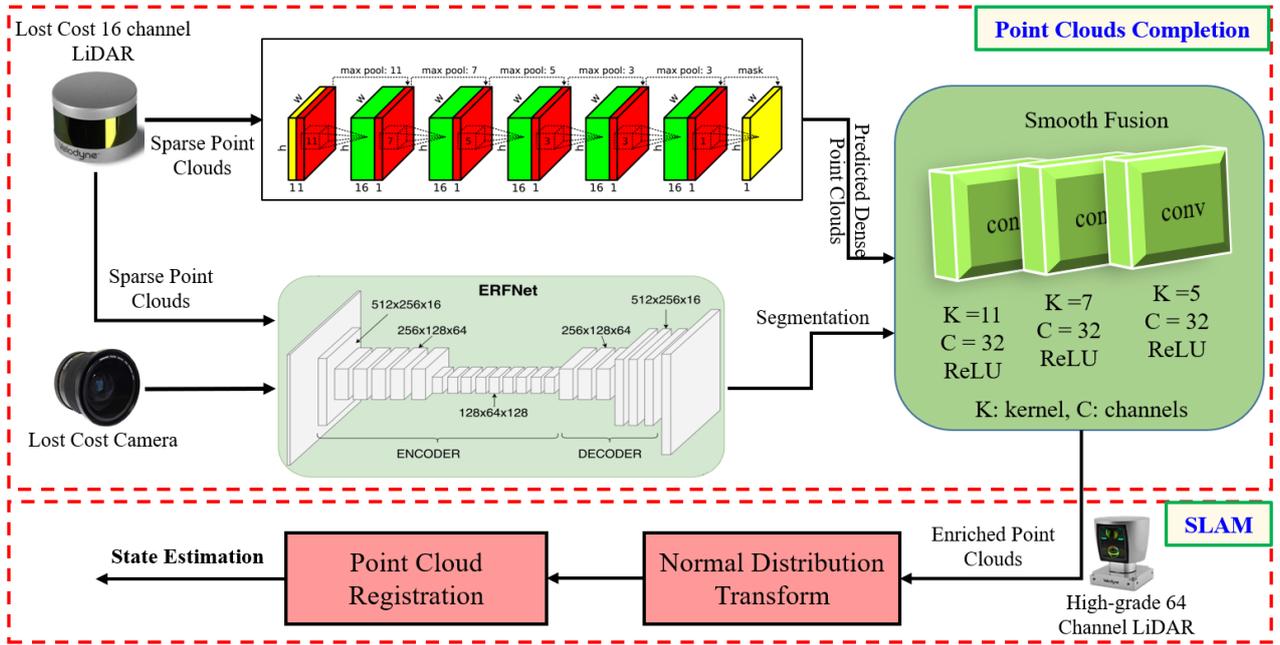

Fig. 1. The framework of the proposed method. The inputs are the 3D point clouds from 16 channels of LiDAR and colored images from a camera. The output is the dense 3D point clouds which have similar density with the 64 channels of LiDAR.

The major contributions of this paper are listed as follows:

(1) This paper conducted a preliminary study on the sparsity and target detection of the depth data. Based on the conclusions, a new neural network is proposed to enrich the simulated 16-channels LiDAR up to 64-channels.

(2) In addition to the depth enrichment results, we benchmarked the SLAM on the generated dense 3D point clouds and 64 channels LiDAR data (ground truth). It is found out that the SLAM result is significantly denser without introducing large error.

(3) This paper evaluated the sparsity of the depth, concluded that the depth is sparse and depth is better on target segmentation. Inspired by these two conclusions, the LiDAR enrichment algorithm could be having better results, if we applied the depth input into the segmentation.

IV. METHODOLOGY

Different from traditional pattern recognition problems, the depth completion is more similar to a measurement problem. It belongs to the so-called ill-posed problem [20]. We will first evaluate the sparsity of the depth and advantage of edge detection. Then, we show that based on the state-of-the-art depth completion method, the data enrichment using 16 channel LiDAR can achieve a performance that is similar to that using 64 channel one. Due to the advantage of the implmented target segmentation of depth, we have better results while the segmentation is aided by the depth. Thus, we demonstrated two important conclusions, the depth is sparse for most of the scene, and the depth aided segmentation would improve the LiDAR point cloud enrichment.

A. Preliminary Study on the Sparsity of Depth

As we mentioned earlier, the sparse is important for depth data completion. If the data is sparse, it means there is a lot of redundancy data in the depth map. We could realize a data enrichment with LiDAR input only. In contrast, if the data is not sparse, then extra input should be needed to keep the results of enrichment reliable. Discrete cosine transform (DCT) is employed to perform the results of the depth map in different compressed rate. An example is illustrated in Fig. 2.

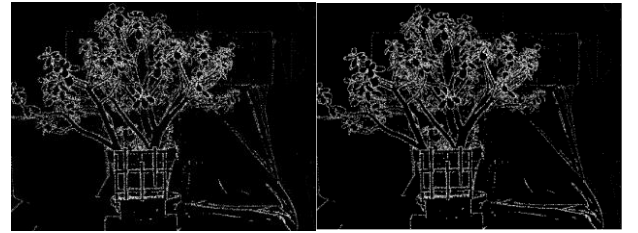

Fig. 2. The discontinuity area of different compressed rate of the depth map. (left image) Compressed rate=1 means there is no compress. (right image) If we keep only 0.0625 information of the ground truth, the depth is still well reconstructed.

From Fig.2, we can see that even in a highly compressed rate, the discontinuity of the depth map still could be well retrieved. It provides a good probability to deal with the targets. On the other hand, a general understanding in the depth completion is that scenes with semantically similar appearances should have similar depth distributions [17]. It is important that if we can detect a real edge to produce a good segmentation. Normally, segment a target with texture is an ill prose-problem, we can't detect a true edge sometimes. But if the point cloud is introduced the result will be different.

The target segmentation and edge detection with the image are two similar tasks, relying on extra priors to estimate accurate results. But the depth map has a natural advantage for target segmentation and edge detection, compared with the image. We analyzed the discontinuity of the targets, both the image and point cloud. As we pointed earlier, the discontinuity of the depth is the critical problem for LiDAR enrichment, which caused the uncertainty of the boundaries of targets. Fig.3 shows the relationship between the image edge and the discontinuity of the surface.

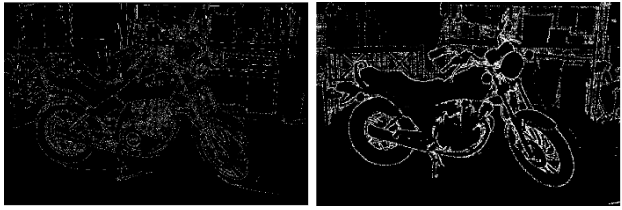

Fig. 3. Edge of image (left image) and discontinuity of depth image (right image). The edges extracted from the image are very ambiguous, hard to provide a useful target prior to LiDAR enrichment.

As we can see in Fig.3, the error mainly caused by the discontinuity of the targets. However, the existing methods normally treat the surface as a continuous problem. Also, the data enrichment methods embedded filters are based on a simple assumption that the areas with the same texture have the same depth. Unfortunately, none of these methods could deal with the discontinuous of targets.

*B. Observation Matrix*

According to the sparsity invariant CNN, an observation mask is added to render the filter output invariant to the actual number of observed inputs. The following figure shows the observation matrix, called sparse convolution.

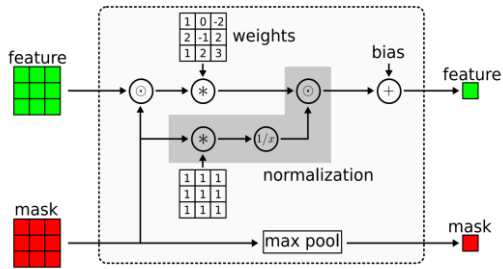

Fig. 4. The observation matrix of the input sparse LiDAR point cloud data, which keeps each convolutional layer weight the input point only.

Different from the traditional convolution kernel, the sparse convolution operation can be written as

$$f_{u,v}(x, o) = \frac{\sum_{i,j=-k}^{k} o_{u+i,v+j} x_{u+i,v+j} w_{i,j}}{\sum_{i,j=-k}^{k} o_{u+i,v+j} + \epsilon} + b \quad (1)$$

where the $O_{u+i,v+j}$ is the pixel in the observe matrix, the $x_{u+i,v+j}$ is the pixel in the depth map and the $W_{i,j}$ is the weight to be trained.

Here, we employed a 2D convolution network to extract the feature of the high-resolution image to segment the targets in the scene. Due to the lack of detail, the sparse depth data cannot retrieve the target well. Fortunately, the image could provide the missing details within the target. In this paper, we use four convolution layers in the image branch, which we try to learn the prior of the target feature to retrieve the detail of the depth map.

The network architecture is mainly constructed by four conventional kernels, which are inspired by the VGG16 [21]. Rectified linear units (ReLU) are used as activation functions. This branch is trying to extract the local feature to up-sampling the sparse depth map.

TABLE I. THE NETWORK OF IMAGE AND SPARSE CONVOLUTION OUTPUT

| Layer | Kernel | Filters |
|---|---|---|
| Conv/Relu | 11x11/1 | 32 |
| Conv/Relu | 7x7/1 | 32 |
| Conv/ Relu | 5x5/1 | 32 |
| Conv/ Relu | 1x1/1 | 32 |

*C. Loss Function*

Since we turn the depth map into an image, we use the L2 as the loss function to keep the smooth of the result. However, the L1 is robust at disparity discontinuities and has low sensitivity to outliers or noises [22]. Another penalty terms are added, which the equation is shown as follows

$$L = \sum_{i \in P} \left( \left\| D_i^{gt} - D_i \right\|^2 + \lambda \left\| D_i^{gt} - D_i \right\|^1 \right) \quad (2)$$

where $\lambda$ is the regularization parameter, which is given as 0.0002 in this paper.

*D. LiDAR SLAM baesd on Normal Distribution Transform*

The LiDAR SLAM consists of two parts, the front end [23], and the back end [23]. The front end focuses on the LiDAR odometry and loop closure [24]. The back end performs the optimization by integrating the constraints from LiDAR odometry and loop closure.

The principle of LiDAR odometry [25] is to track the motion differences between two successive frames of 3D point clouds by matching the two frames (called as a reference and an input point cloud in this paper). The matching process is also called point cloud registration. The objective of point cloud registration is to obtain the optimal transformation matrix to match or align the reference and the input point clouds. The most well-known method of point cloud registration is the iterative closest point (ICP) [26]. The ICP is a straightforward method to calculate the transformation matrix between two consecutive scans by iteratively searching pairs of nearby points in the two scans and minimizing the sum of all point-to-point distances. The objective function can be expressed as follows [26]:

$$C(\hat{R}, \hat{T}) = \arg\min \sum_{i=1}^{N} \|(Rp_i + T) - q_i\|^2 \quad (3)$$

where the $N$ indicates the number of points in one scan $p$, $R$ and $T$ indicate the rotation and translation matrix, respectively, to transform the input point cloud ($p$) into the reference point cloud ($q$). The objective function $C(\hat{R}, \hat{T})$ indicates the error of the transformation. One of the main drawbacks of this method is that ICP can easily get into the local minimum problem. The normal distribution transform

[27] (NDT) is a state-of-art method to align two consecutive scans with modeling of points based on Gaussian distribution. The NDT innovatively divides the point cloud's space into cells. Each cell is continuously modeled by a Gaussian distribution. In this case, the discrete point clouds are transformed into successive continuous functions. In this paper, the NDT is employed as the point cloud registration method for the LiDAR SLAM. Assuming that the transformation between two consecutive frames of point clouds can be expressed as $T = [t_x\ t_y\ t_z\ \phi_x\ \phi_y\ \phi_z]^T$. The $t_i$ indicates the translation in the $x$, $y$, and $z$-axis, respectively. The $\phi_x$ represents the orientation angle of the roll, pitch, and yaw, respectively. Steps of estimating the relative pose between the reference and the input point clouds are as follows:

(1) Fetch all the points $x_{i=1\ldots n}$ contained in a 3D cell [28].
Calculate the geometry mean $q = \frac{1}{n}\sum_i x_i$.
Calculate the covariance matrix

$$\Sigma = \frac{1}{n}\sum_i (x_i - q)(x_i - q)^T \quad (4)$$

(2) The matching score is modeled as:

$$f(p) = \sum_i \exp(-\frac{(x_i' - q_i)^T \Sigma_i^{-1} (x_i' - q_i)}{2}) \quad (5)$$

where $x_i$ indicates the points in the current frame of scan $p$. $x_i'$ denotes the point in the previous scan mapped from the current frame using the $T$. $q_i$ and $\Sigma_i$ indicate the mean and the covariance of the corresponding normal distribution to point $x_i'$ in the NDT of the previous scan.

(3) Update the pose using the Quasi-Newton method based on the objective function to minimize the score, $f(p)$.

With all the points in one frame of point clouds being modeled as cells, the objective of the optimization for NDT is to match current cells into the previous cells with the highest probability. The optimization function $f(p)$ can be found in [27]. Therefore, $T$ can be estimated by optimization.

## V. EXPERIMENT EVALUATION

### A. Experiment Setup

To verify the effectiveness of the proposed method, both the 3D point cloud enrichment and SLAM are evaluated. To quantitatively evaluate the results, both mean absolute error (MAE) and root mean squared error (RMSE) are employed, which can be written as:

$$MAE = \frac{\sum_{i=1}^{n} |y_i - y_i|}{n}$$

$$RMSE = \frac{\sqrt{\sum_{i=1}^{n} (y_i - y_i)^2}}{n} \quad (6)$$

where $y_i$ is the ground truth of the depth, $y_i$ is the predicted depth, $n$ is the total number of the point clouds.

Regarding the data enrichment, there are three benchmarks are presented:

(1) **L2L**: LiDAR data input only to enrich the 3D point clouds [13].
(2) **LI2L**: LiDAR and image data are inputted to the network independently [18].
(3) **F2L**: LiDAR aided image segmentation plus LiDAR based on Fig 1, which is the proposed method.

### B. Train Dataset

For the experiments, a 2080Ti GPU was used and the code is implemented in Pytorch [29]. We evaluate our framework by computing the loss on all pixels of the ground truth since not all input pixels of the LiDAR are correct. There are 1730 training frames were selected (7 days acquisition from KITTI), which is downsampled from 64 channels LiDAR data to simulate the 16 channels LiDAR. We selected 154 (1-day acquisition) frames as the test data.

### C. Evaluation of Data Enrichment

Table II shows the results of the data enrichment. The second column shows the RMSE. The third column shows the MAE. 2.165 meters of MAE is obtained using the L2L method. After integrating the image with the LiDAR data during the enrichment based on [18], the error even increases to 2.622 meters. This is due to the lack of well pre-trained image sementation model, prior of segmentation. Actually, even the entire KITTI dataset could not provide the prior, extra dataset needed [12, 18]. With the help of the proposed method, the MAE decreases to 1.153 meters which shows the effectiveness of the proposed method.

TABLE II. THE PERFORMANCE OF THE DATA ENRICHMENT

| Method | RMSE (mm) | MAE (mm) |
|---|---|---|
| LiDAR to LiDAR (L2L) [13] | 22160 | 2165 |
| LiDAR/image to LiDAR (LI2L) [18] | 26162 | 2622 |
| fusion to LiDAR (F2L) | 8339 | 1153 |

Fig. 5 shows the results of the data enrichment. Fig. 5 (a) shows the simulated sparse 16 channels LiDAR point clouds. With the help of the proposed method, the density of the point clouds is significantly enhanced which is shown in Fig. 5 (b). The bottom figure shows the ground truth point cloud data from 64 channels LiDAR. We can see that the proposed method in this paper obtains comparable density with the ground truth.

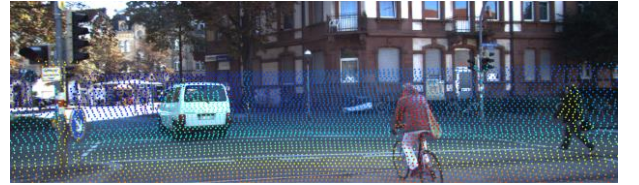

(a) Simulated 16 channels LiDAR input

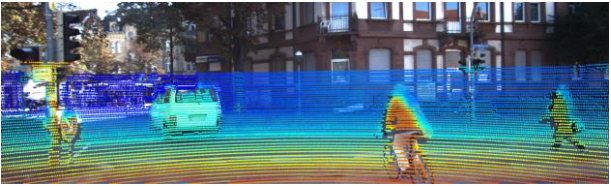

(b) Enriched result up to 64 channels LiDAR by our proposed method

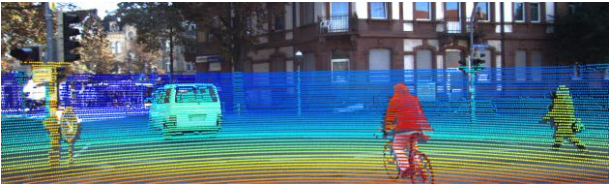

(c) 64 channels LiDAR result (Ground truth)

Fig. 5. The LiDAR enrichment result and ground truth. (a) simulated 16 channels input; (b) enrichment result; (c) the ground truth from 64 channels LiDAR.

### D. Evaluation of SLAM based on the Enriched 3D Point Clouds

Furthermore, we benchmarked the predicted 64 channels LiDAR data into SLAM. The results are shown in the following figure.

TABLE III. THE PERFORMANCE COMPARISON OF SLAM USING THE SIMULATED LiDAR 16 CHANNELS POINT CLOUDS AND THE LiDAR POINT CLOUDS ENRICHED ENRICHED BY THE PROPOSED METHOD.

| All data | Simulated 16 Channels (m) | Proposed Method (m) |
|---|---|---|
| Max error | 0.784 | 1.200 |
| Mean error | 0.247 | 0.250 |
| Min error | 0.025 | 0.024 |
| RMSE | 0.295 | 0.371 |
| STD | 0.161 | 0.274 |

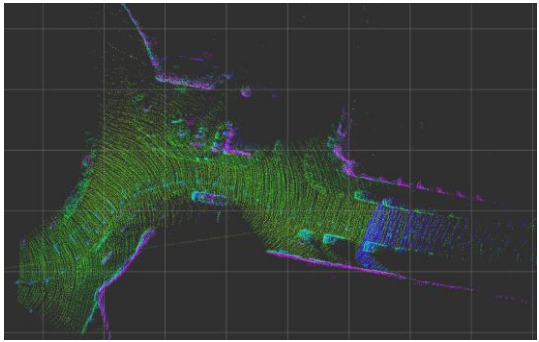

(a) SLAM result based on simulated 16 channels LiDAR, MAE 0.24m

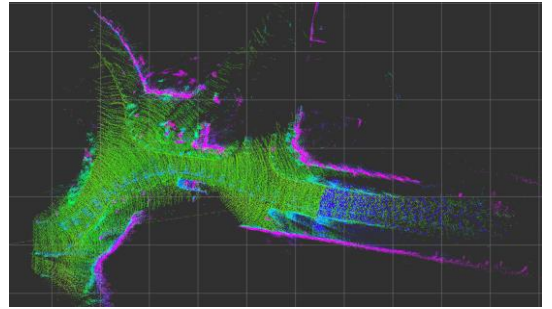

(b) SLAM result based on the proposed LiDAR enrichment neuronal network (64 channels), MAE 0.25m

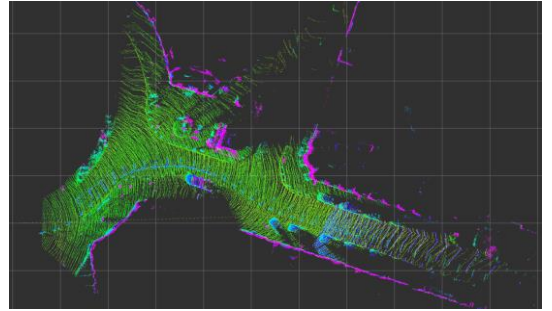

(c) SLAM result based on 64 channels LiDAR (Ground truth)

Fig. 6. The SLAM results based on simulated 16 channels LiDAR, 64 channels LiDAR by enrichment, 64 channels LiDAR ground truth.

From the results of SLAM, we can see that the density of the maps from SR (Fig.6 (b)) and the ground truth (Fig.6 (c)) is comparably the same. But we obtain almost the same accuracy of the SLAM with the 16 channels LiDAR (Fig.6 (a)). The reason of the lack of improvement in accuracy could be that the training dataset applied in this paper is still small. In the near future, the SLAM result would be improved if a larger training dataset is applied. Based on the proposed LiDAR data enrichment (LiDAR depth completion), we show the SLAM using low-cost LiDAR (16 channels one) after enrichment can achieve a localization performance that very similar to that of a high-performance LiDAR (64 channels one).

In fact, the LiDAR SLAM strongly affected by the dynamic objects, which makes it not satisfactory for autonomous driving applications. No double, there are two significant issues; highly dynamic road caused a lot of 1) occlusions and 2) scene changing, which affects both LiDAR depth completion and LiDAR SLAM. Our future work will focus on the development of a new LiDAR depth completion method to mitigate the above-mentioned issues.

## VI. CONCLUSIONS AND FUTURE WORK

The sparsity of data is an important characteristic. The realization of data compression and enrichment depends on the sparsity. In this paper, we show that the depth could be well retrieved in a heavily compressed condition, which provided a guide to design the LiDAR data enrichment neuronal network. Since the traditional LiDAR enrichment algorithms could have good results with different filters, which means that similar texture has similar depth [30]. But how to segment the image into the right targets would be another big problem. We compared the edge extracting results on sparse depth data and images, which show that the depth has much better results. This conclusion inspired us to employ the sparse input data together with the image to achieve better target segmentation.

We show that with the help of segmentation of the depth, the proposed LiDAR data enrichment neural network has gained significant improvement on our tested KITTI dataset. We also benchmarked the performance of LiDAR SLAM on our estimated 64 channels LiDAR data which have obtained denser point cloud map than the one from the original simulated 16 channels LiDAR input.

However, the simulated 16 channels LiDAR input used in this paper is average pooling from the 64 channels LiDAR results, the simulated data is different from a real 16 channels LiDAR. On the other hand, the traffic in the KITTI dataset is common quite different from the highly urban area. In future work, we will collect challenging urban on-road datasets in Hong Kong to verify the effectiveness of the proposed method. Moreover, we will the potential of the denser 3D point clouds generated by the proposed method to help GNSS positioning, for example, detecting [31, 32] the GNSS NLOS or correcting [33, 34] the GNSS NLOS measurements.

ACKNOWLEDGMENT

The authors acknowledge the support of the Hong Kong PolyU internal grant on the project ZVKZ, "Navigation for Autonomous Driving Vehicle using Sensor Integration".